\documentclass{article}

\usepackage[english]{babel}

\usepackage[letterpaper,top=2cm,bottom=2cm,left=3cm,right=3cm,marginparwidth=1.75cm]{geometry}

\usepackage{amsmath}
\usepackage{graphicx}
\usepackage[colorlinks=true, allcolors=blue]{hyperref}

\usepackage{appendix}
\usepackage{authblk}

\usepackage[
backend=biber,
style=numeric,
sorting=ynt
]{biblatex}

\usepackage{booktabs}
\usepackage{multirow}
\usepackage{csquotes}

\title{A Survey of Pipeline Tools for Data Engineering}
\author{Anthony Mbata,~Yaji Sripada,~and~Mingjun Zhong}
\affil{Department of Computing Science, University of Aberdeen, UK}
\affil{\{a.mbata.22,yaji.sripada,mingjun.zhong\}@abdn.ac.uk}

\begin{document}
\maketitle

\begin{abstract}
Currently, a variety of pipeline tools are available for use in data engineering. Data scientists can use these tools to resolve data wrangling issues associated with data and accomplish some data engineering tasks from data ingestion through data preparation to utilization as input for machine learning (ML). Some of these tools have essential built-in components or can be combined with other tools to perform desired data engineering operations. While some tools are wholly or partly commercial, several open-source tools are available to perform expert-level data engineering tasks. This survey examines the broad categories and examples of pipeline tools based on their design and data engineering intentions. These categories are Extract Transform Load/Extract Load Transform (ETL/ELT), pipelines for Data Integration, Ingestion, and Transformation, Data Pipeline Orchestration and Workflow Management, and Machine Learning Pipelines. The survey also provides a broad outline of the utilization with examples within these broad groups and finally, a discussion is presented with case studies indicating the usage of pipeline tools for data engineering. The studies present some first-user application experiences with sample data, some complexities of the applied pipeline, and a summary note of approaches to using these tools to prepare data for machine learning.
\end{abstract}

\section{Introduction}

\label{sec:introduction}

The vast amount of available data and data generated now in real-time is associated with \emph{data wrangling} or \emph{data engineering} \cite{nazabal2020data}, namely obtaining, organising, understanding, extracting, and formatting data for analysis. The purpose of data engineering is to manipulate the raw data to structured in a desired form so that the data can be used as input to downstream tasks like machine learning models. These tasks of data engineering are tedious and time-consuming and thus would consume most of the data scientist's time in a data science or artificial intelligence (AI) project; for example, it was estimated that a data scientist would spend $80\%$ of the time on data engineering in a data science project \cite{dasu2003exploratory}. 

The challenge data engineering faces is that there is no formula to transform raw data into the desire form automatically. For tackling such challenge, there have been different levels of advancements developed by corporate entities and academia. The key is to implement data engineering procedure as a series of semi-automated or automated operations to transform raw data into the structured data as required. Such procedures are viewed as data engineering pipeline frameworks \cite{widanage2020high,De_Bie_2022,Harald2023}. 

This survey aims to identify and review those pipeline tools and frameworks which implement data processing procedures attempting to resolve data engineering challenges. A variety of tools and frameworks exist, and each focuses on different tasks of data engineering. For example, some tools are customized to handle those data engineering issues posed by a given dataset; others focus on data engineer's understanding and engagement with a tool or framework and its extensibility and integration with other tools. A one-size-fits-all framework is ideal, but would not be possible as the framework would need to be customized for various datasets. Indeed, a more realistic framework should have the functionality for users to add custom solutions to the existing tools as new data might produce a new challenge and require custom solutions. Where a tool can integrate seamlessly with other tools, it provides a better opportunity for resolving data engineering issues and is useful for creating pipelines for individual solutions within a framework. 

In this paper, we present a survey of pipeline tools and frameworks for data engineering that are available for providing solutions to broad data engineering issues and examples of implementing solutions with these tools on a given dataset. These tools or frameworks have been grouped into categories based on their overall design and use cases. While some frameworks or platforms can fall into more than one category, their grouping is broadly based on some common features and functionality. One of these groups is Extract Transform Load (ETL) / Extract Load Transform (ELT) pipelines, designed to provide mainly data integration solutions. Secondly, pipelines for Data Integration, Ingestion, and Transformation are designed to provide solutions associated with data organisation, for example, organising data from multiple sources. Thirdly, Orchestration and Workflow Management pipelines are designed to automatically organise and manage pipeline workflows. Finally, Machine Learning Pipelines are specifically designed for machine learning and model deployment into production. While one size may not fit all, the categorisation of pipelines that are useful for data engineering pipelines aids the choice of solution a data engineer has in resolving the multiple challenges associated with data, as well as provide a guide to research into the improvement of existing solutions.

In addition, the paper provides a review of existing pipelines vis-a-vis a dataset with some issues, i.e., the IDEAL Household Energy Dataset \cite{pullinger2021ideal}. Some experiments present certain features and functionality of a few data pipelines, showing how they resolved some data organisation, data quality, and feature engineering issues found in the IDEAL Household Energy dataset to help prepare the dataset and make it ready as input for machine learning.

The paper is organised as follows: Section \ref{Sec2} introduces generalised data engineering pipelines and the essence of designing them. In Section \ref{Sec3} we describe the survey methodologies for those pipeline tools. Section \ref{Sec4} discusses the different groups of pipelines based on their functions. In Section \ref{Sec5} case studies are presented for selected pipelines. Section \ref{Sec6} discusses trends in the features of the selected pipelines, some recommendations are provided in Section \ref{Sec7}, and our conclusion is presented in Section \ref{Sec8}.

\section{Data Engineering Pipelines}
\label{Sec2}

The combination of some steps and processes from the data source to the destination where the data is utilized makes up a data pipeline. For example, in a data science project, the data must be in a specific form before it is fed into a machine learning model. Appropriate data pipelines may be applied to transform that data into that specific form. The consumption of fit-for-purpose data requires that data is sourced, flows to, and becomes available where it is consumed. It takes the important processes of capturing and ingesting this data into the pipeline, integrating the various sourced data, and preparing the data for use. These require coordination through a well-organised data orchestration and timely workflow management of the different processes and their seamless connectivity across and along the pipeline ensuring that the data flows through and is prepared to a form fit for intended utilisation. In designing an effective and efficient data pipeline framework, many questions come into mind such as what type of data and its format, can the data be read, and if yes, how can the data be organised? With data quality concerns can the data be cleaned also without missing data? Regarding engineering of features to improve the usefulness of the data, can this be done? The design of pipelines takes a logical process, and can take the form of an ETL (Extract Transform Load) where raw data is extracted from sources, then transformed into the required configuration and stored at the destination for consumption. For big data, the format is preferably an ELT (Extract Load Transform) where raw data is extracted and loaded into locations such as data warehouses, and data lakes before transformation. Since data comes from a variety of sources and in different formats, it becomes imperative to perform data integration, i.e., the process of combining multi-source heterogeneous data \cite{De_Bie_2022}. Data integration has consistently evolved and expanded with technological advancement \cite{xiaojuan2023data}. For example, in artificial intelligence and data management, both spheres of advancement have improved data integration \cite{dong2018data} and overall benefited data engineering pipeline frameworks. 

\begin{figure*}
    \raggedright 
    \includegraphics[width=15.50cm, height=6.75cm]{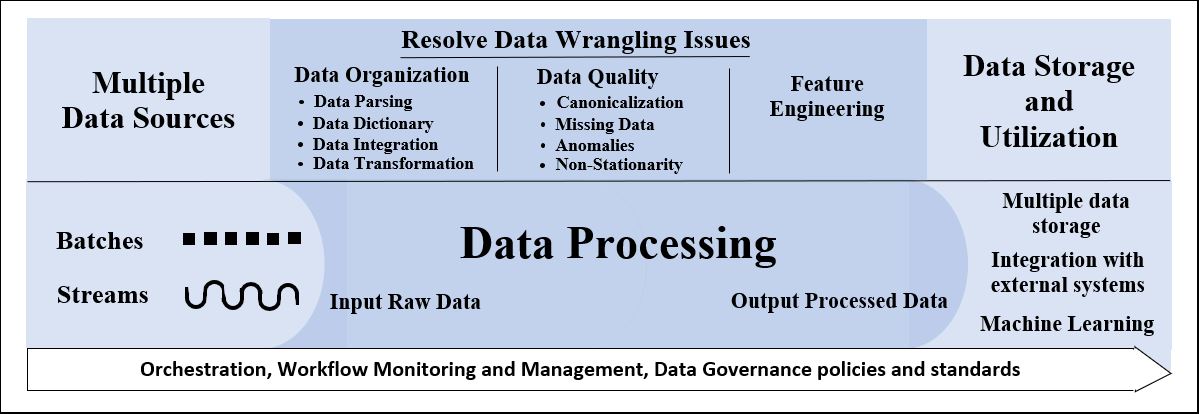}
    \caption{An ideal pipeline for data engineering (adapted from \cite{nazabal2020data}). The schematic points to data sourcing from multiple sources whether sourced in batches or streaming data. The data processing aspects cover the activities that occur between receiving the data and outputting the same for further use. Data processing aims to resolve all the listed data wrangling issues, however, in practice, some are resolved rather than all. Once completed the data is stored for further use or immediately consumed for visualisation and machine learning.}
    \label{fig:ideal_DE_pipeline}
\end{figure*}

In an ideal design and function, the data pipeline should be flexible and able to adapt smoothly to handling any data situation. Ideally, a data pipeline is expected to read raw data inputs in the data ingestion process. Multiple sourced data could be ingested in batches for instance at specific times or triggered when a set condition is attained. The pipeline is also expected to ingest real-time data stream and transmit the same across through the different operations down the system as illustrated in Figure \ref{fig:ideal_DE_pipeline}. There are several moving parts within a data pipeline, for example, a pipeline might have been designed to ingest small data, but a change in the requirement to ingest big data might negatively impact the entire process. An ideal data pipeline is expected to have the ability to deal with an issue like this by automatically scaling up in this example, and when required it should also be able to scale down with ease. Essentially, a change in requirements should not break the pipeline rather the pipeline should be flexible enough to seamlessly accommodate these changes.

The levels of support available for data engineering pipelines vary, from the existence of just a container for solutions to the availability of plug-and-play components and features. Some pipelines do exist where a specific section comes with preloaded and ready-to-use components such as TFX (TensorFlow Extended) pipeline for machine learning, it's however a wish to have an ideal pipeline where the components at every stage of the pipeline are preloaded and scalable to accommodate changing requirements for data engineering tasks.

The volume of the data passing through the pipeline needs to be handled seamlessly irrespective of the speed and rapid change of volume and diverse nature of the data inflow. Whether as batch or streaming data, when received the pipeline should engage the full suite of data processing operations, resolve applicable data issues, and ensure that the data is processed completely. As a minimum requirement when data processing is completed, the pipeline should output the processed data into some form of storage, consumption, integration with other systems, or made available for machine learning. While performing all these the pipeline maintains end-to-end automation and monitoring of the system, all in line with data governance regimes. Such an ideal fit-for-all pipeline for data engineering that solves all data wrangling issues end-to-end would be great to have. Issues still abound regarding the operability of such pipelines, the compatibility within and across components, and with data types \cite{foidl2024data}. Largely pipelines in existence are designed to resolve specific data wrangling issues, and a combination of multiple pipelines into a framework would provide more benefits.

\section{Methodology}
\label{Sec3}
The survey methodology for the tools, pipelines, or frameworks involved an initial scanning of the web to get an overall understanding of the web community about available resources and to have an idea about the discourse from the technical and non-technical community. This provided a general understanding of some names and groupings of pipeline resources. Some were identified as completely open and free for public use and others tend to be closed-sourced tools with some commercial focus. Further information was derived by reviewing pipeline documentation on dedicated websites, some of which were company or commercial websites for closed-source tools. This further reading supplied extra in-depth understanding and information on the progressive development of these pipeline resources such as the dates of introduction, primary supported languages, and some overall information on applicability. 

Insights from developers, sales pitches, and promotions from commercial entities supplied some technical and commercial competitiveness of the use of some pipelines for data engineering tasks. Depending on the intended use, some of this information explained how tools/pipelines can be designed and utilised by non-technical and technical users to achieve data organisation, data quality, and feature engineering. Some of these possibilities require prior technical knowledge and depend also on tool built-in capabilities, customisation, or combination with other tools to perform these data engineering tasks.

More technical-focused surveys of these pipeline tools were achieved by investigating the GitHub repositories that have been designed for these tools. User communities as well as dedicated users and projects helped to garner some understanding of the pipeline ecosystem. Moving further to more technical understanding, technical publications provided a better insight into some formal applications and discussions within the technical community. As the main focus of the survey was to look at what exists within the data engineering pipeline space, the intention was to strike a balance to allow even non-technical audiences to have an understanding of available pipelines for data engineering as well as find it easy to engage in some hands-on experience with a tool in an attempt to perform some data engineering task on an example dataset. 

At the time over 40 different tool materials, websites, promotion materials, documentation, and technical papers were reviewed, some of these were simply for information and awareness. However, four of these pipelines were used to perform some basic data engineering tasks to demonstrate some of their capabilities and functionality using mostly their inbuilt features. The decision to choose these four as representatives for actual practical engagement was based on whether the tool is open source (O) or close source (C), ease of use, and some capabilities described from technical, commercial, and public user experiences. These representatives are \textit{Apache Spark}, \textit{Microsoft SQL Server Integration Services}, \textit{Apache Airflow}, and \textit{TFX (TensorFlow Extended)} respectively.

These four fall within broad data engineering pipeline categories: (a) ETL/ELT Pipelines, (b) Data Integration and Ingestion, and Data Transformation, (c) Pipeline Orchestration and Workflow Management, and (d) Machine Learning and Model Deployment. Some pipeline tools can be in more than one group, also there could be variations in capabilities within a specific grouping but for practical hands-on engagement just one tool was chosen per group for the survey. The overall intention is to provide a generalisation of how data engineering pipelines can be designed and to assist technical and non-data technical users in achieving some data engineering tasks. The survey is not to indicate a comparative analysis of these pipelines but a general survey of their availability, some of their broad usability, and some first user experience which is provided as methods in Section 5.1. 

\section{Clusters of Data Pipelines}
\label{Sec4}
Data pipelines could be grouped into some categories depending on the factors under consideration. A batch pipeline processes data from source to destination at certain predefined times or conditions. Streaming data pipelines is an improvement to batch processing \cite{hlupic2021overview}. Here data is processed in real-time from source to destination as the input data is received, this is necessary to fulfill the demands for quick information and results. With rapid information and data production, the need for scalability leads to increasing cloud operations as such data pipelines could be grouped based on the operating environment whether local or on-premises types and pipelines that can be deployed for cloud-based operations. The grouping of pipelines can also be considered across the key functions for which they have been designed. 
These pipeline clusters/groups are 
\begin{itemize}
  \item ETL/ELT Data Pipelines
  \item Integration, Ingestion, and Transformation
  \item Pipeline Orchestration \& Workflow Management
  \item Machine Learning and Model Deployment
\end{itemize}

\subsection{ETL/ELT Data Pipeline}
ETL pipelines are designed to perform an aspect of data integration through the extraction of data from sources. Then the data is transformed before loading into the destination. Big data counterparts integrate data using an ELT process where the pipelines also extract data but load it into a destination before the transformation process. There are suites of ETL/ELT pipeline tools and frameworks available as listed in Table \ref{tab:ETL_/_ELT_Pipelines}. The table gives a summary of the year some of these tools/frameworks/platforms were introduced, whether these are open sources (O) or closed sources (C), the primary languages of these tools, and their capabilities in terms of built-in (B) or can integrate (I) with other tools to accomplish the said capabilities.

\textit{Apache Spark} is an open-source platform that supports multiple languages Python, Java, SQL, Scala, and R, and executes data engineering, data science, and machine learning tasks. It provides a platform for distributed and scalable large-scale parallel data processing suitable for quick big data query and analysis, provides built-in capabilities and useful applications to feature engineering in data wrangling \cite{van2020evaluation}, analysing different forms of data including graph processing \cite{aseman2022cost} referred to as Directed Acyclic Graph (DAG).
Spark integrates and processes data in both local and cloud environments, with web-based applications in batch and real-time mode. The parallel processing in Apache Spark involves the processing of several complex tasks simultaneously, coupled with its flexibility, connected graphs (DAG) handling of tasks, in addition to other capabilities gives some advantages of using Apache Spark-based pipelines. However, when these graphs become too long the reliability of Apache Spark pipelines is affected \cite{kweun2020lineage} as it becomes failure-prone with an overall negative effect on the performance.

\textit{AWS Glue} is a closed source Serverless ETL, ELT service, that helps to simplify, monitor and manage data pipelines (https://aws.amazon.com/glue/). It connects and integrates with other tools for machine learning, analytics, and app development, and supports multiple languages Apache Spark, Ray, and Python backends. Glue supports multiple frameworks and workflow solutions in batch and streaming data mode, functions with a visually interactive interface, supports codeless functions, and connects with notebooks and an integrated development environment (IDE) \cite{farki2023real}. Data engineering issues associated with data quality could benefit from AWS Glue outputs of data quality metrics, monitoring, and management. These same data quality capabilities exist for \textit{Apache NiFi}, \textit{Striim} which in addition has built-in artificial intelligence (AI) abilities (https://www.striim.com/
), and \textit{SAS Data Integration Studio} an intuitive visual collaborative interface for data integration (https://support.sas.com/en/software/data-integration-studio-support.html).

\begin{table*}

    \renewcommand{\arraystretch}{1.0}
    \setlength{\tabcolsep}{3pt}
    \smallskip
    \caption{ETL / ELT Pipelines}
    \label{tab:ETL_/_ELT_Pipelines}
    \begin{tabular}{|l|c|c|c|*{4}{p{0.3cm}|}*{4}{p{0.3cm}|}c|*{3}{p{0.3cm}|}c|*{2}{p{0.3cm}|}l|}
    \toprule
    \multirow{2}{*}{Tools/Framework/Platform} &
    \multicolumn{1}{c|}{\rotatebox{90}{Year of Introduction}} &  
    \multicolumn{1}{c|}{\rotatebox{90}{Source}} &
    \multicolumn{1}{c|}{\rotatebox{90}{Primary Language}} &
    \multicolumn{4}{c|}{\rotatebox{90}{Data Organisation}} & 
    \multicolumn{4}{c|}{\rotatebox{90}{Data Quality}} &
    \multicolumn{1}{c|}{} &
    \multicolumn{3}{c|}{\rotatebox{90}{Processing Format}} &
    \multicolumn{1}{c|}{} &
    \multicolumn{2}{c|}{\rotatebox{90}{Environment}} \\
    \cline{5-18}
    &&&& \rotatebox{90}{Data Parsing} & \rotatebox{90}{Data Dictionary} & \rotatebox{90}{Data Integration} & \rotatebox{90}{Data Transformation} & \rotatebox{90}{Canonicalization} & \rotatebox{90}{Missing Data} & \rotatebox{90}{Anomalies} & \rotatebox{90}{Non-Stationarity} & \rotatebox{90}{Feature Engineering} & \rotatebox{90}{ETL} & \rotatebox{90}{Batch} & \rotatebox{90}{Streaming} & \rotatebox{90}{DAG / Graph} & \rotatebox{90}{Local} & \rotatebox{90}{Cloud} \\
    \hline
    \\
    Apache Spark                & 2010 & O & Scala, Java     & B & I & B & I & I & I & I & I & B & I & B & B & B & B & B \\
    Apache Flink                & 2010 & O & Java            & I & I & I & I & I & I & I & I & I & I & I & B & B & B & B \\
    Apache Griffin              & 2016 & O & Scala           & I & I & I & I & B & B & B & B &   & I &   &   & I & B & B \\
    Luigi                       & 2012 & O & Python          & I & I & I & I & I & I & I & I & I & B & B &   & B & B & I \\
    Dagster                     & 2019 & O & Python          & I & I & I & I & I & I & I & I & I & I & I & I & B & B & B \\
    Kedro                       & 2019 & O & Python          & I & I & I & I & I & I & I & I & I & B & B & I & B & B & I \\
    DBT (Data Build Tool)       & 2016 & O & SQL(Jj)         &   &   & I & B & I &   &   &   &   & B & I &   & B & B & I \\
    Bonobo                      & 2013 & O & Python          &   &   & I &   & I &   &   &   &   & B &   &   & B & B &   \\
    AWS Glue                    & 2017 & C & Python, Scala   & B &   & B & B & I &   &   &   &   & B & B & I & B &   & B \\
    Snowflake Data Pipelines    & 2014 & C & SQL             &   &   & I & B & I &   &   &   &   & B & I &   & I &   & B \\
    Google Cloud Dataflow       & 2015 & C & Java, Python    &   &   & B &   & I &   &   &   &   & B & B & B & B &   & B \\
    Databricks Delta Lake       & 2017 & O & Scala, Python   &   &   & I &   & I &   &   &   &   & B & I &   & B & B & B \\
    StreamSets                  & 2015 & O & Java, Python    &   &   & I &   & I &   &   &   &   & B & B & B & B &   & B \\
    Dataiku                     & 2013 & C & Python, R       &   & B & I &   & I & I &   &   &   & B & I &   & B & B & B \\
    Alteryx                     & 2010 & C & Py, R, SQL, Ax  &   & B & I &   & I &   &   &   &   & B &   &   & B &   & B \\
    SAS Data Integration Studio & 2005 & C & SAS             & B &   &   &   &   &   &   &   &   &   & B &   & B & B &   \\
    Striim                      & 2012 & C & SQL, XML, JSON, JS        &   &   & B &   &   &   &   &   &   &   &   &   & B &   & B \\
    
   \hline
   
   \end{tabular}
   \\
        
    a. Py: Python; Jj: with Jinja templating; JS: JavaScript; Ax: Alteryx Expression Language \\Within the capabilities, B: Built-in capabilities; I: Integrates with other tools to achieve.

\end{table*}

\subsection{Integration, Ingestion, and Transformation} Data integration, ingestion, and transformation pipelines as shown in Table \ref{tab:Int_Inj_Trfm_Pipelines}, aim to provide solutions to the data engineering issues associated with data organisation for example working with data from multiple sources, heterogeneous data, integrating aspects of data to suite specific requirement. The need to meet these requires that the right data is obtained, integrated, and made available for use, necessitating the importance of having pipelines to accommodate data governance and data management concerns \cite{nadal2022operationalizing}. With the availability of large, complex, and heterogeneous data from multiple sources, the utility of data integration, ingestion, and transformation pipelines is central to data consolidation and transformation. 

\textit{Apache Kafka} is an open-source platform that supports multiple languages Java, Python, Scala, and Go (https://kafka.apache.org/). It can ingest multiple sourced data and transform data during processing \cite{peddireddy2023streamlining}. It reads, writes, and processes data in a local or cloud environment \cite{vyas2022performance} with data security and fault tolerance in place, and scalable in the event of real-time increasing data processing demands and operation. It processes data in batches as well as a platform for high-performance distributed event streaming - processing stream of events in real-time. In terms of speed of data processing, Apache Kafka provides high-speed data processing with low data latency \cite{van2020evaluation}, while monitoring and optimizing operations to prevent failures and ensure seamless operations. Apache Kafka integrates and extends to other tools leveraging this extensibility to do more and has seen more user adoption and engagement for data integration and analytics, providing the advantages of reliable, flexible, efficient, scalable, fault-tolerant, real-time data processing. However, these come with a steep learning curve for the platform and the complexities of setting up, operational monitoring, and support.

\textit{Microsoft SQL Server Integration Services (SSIS)} is a closed-source platform (with open source versions) for building ETL, data integration, and transformation pipeline workflows, and supports SQL and .NET languages (https://learn.microsoft.com/en-us/sql/integration-
services). Microsoft SSIS can ingest multiple files and relational data sources, and transform and load the same into multiple destinations. It runs on-premises and can integrate with the cloud environment, in addition to its graphical interface it can be customised, and easy to use with or without codes and can extend to other tools for increased functionality.

\begin{table*}[!t]

    \renewcommand{\arraystretch}{1.0}
    \setlength{\tabcolsep}{3pt}

    \smallskip
    \caption{Integration, Ingestion, and Transformation}
    \label{tab:Int_Inj_Trfm_Pipelines}
    \begin{tabular}{|l|c|c|c|*{4}{p{0.3cm}|}*{4}{p{0.3cm}|}c|*{3}{p{0.3cm}|}c|*{2}{p{0.3cm}|}l|}
    \toprule
    \multirow{2}{*}{Tools/Framework/Platform} &
    \multicolumn{1}{c|}{\rotatebox{90}{Year of Introduction}} &  
    \multicolumn{1}{c|}{\rotatebox{90}{Source}} &
    \multicolumn{1}{c|}{\rotatebox{90}{Primary Language}} &
    \multicolumn{4}{c|}{\rotatebox{90}{Data Organisation}} & 
    \multicolumn{4}{c|}{\rotatebox{90}{Data Quality}} &
    \multicolumn{1}{c|}{} &
    \multicolumn{3}{c|}{\rotatebox{90}{Processing Format}} &
    \multicolumn{1}{c|}{} &
    \multicolumn{2}{c|}{\rotatebox{90}{Environment}} \\
    \cline{5-18}
    &&&& \rotatebox{90}{Data Parsing} & \rotatebox{90}{Data Dictionary} & \rotatebox{90}{Data Integration} & \rotatebox{90}{Data Transformation} & \rotatebox{90}{Canonicalization} & \rotatebox{90}{Missing Data} & \rotatebox{90}{Anomalies} & \rotatebox{90}{Non-Stationarity} & \rotatebox{90}{Feature Engineering} & \rotatebox{90}{ETL} & \rotatebox{90}{Batch} & \rotatebox{90}{Streaming} & \rotatebox{90}{DAG / Graph} & \rotatebox{90}{Local} & \rotatebox{90}{Cloud} \\
    \hline
    \\
    
    Apache   Kafka         & 2011 & O & Sc, Jv, Py, Go   & I & I & B & I & I & I & I & I & I & I & B & B & I & B & B \\
    Apache Gobblin         & 2017 & O & Java             & I & I & B & I & I & I & I & I & I & B & B & I & I & I & B \\
    Singer                 & 2017 & O & Python           &   &   & B &   &   &   &   &   &   & B & B &   & I & B &   \\
    Stitch                 & 2016 & C & SQL              &   &   & B &   &   &   &   &   &   &   &   &   & B &   & B \\
    Segment                & 2012 & C & JS, Py, Jv, JSON &   &   & B &   &   &   &   &   &   &   &   &   & B &   & B \\
    Fivetran               & 2012 & C & SQL              &   &   & B &   &   &   &   &   &   &   &   &   & B &   & B \\
    Apache NiFi                       & 2014 & O & Java  & B &   & B & B &   &   &   &   &   &   & I & B & B & B & I \\
    Talend Data Integration           & 2006 & O & Java  & B &   & B & B & I &   &   &   &   & B & B &   & B & B & B \\
    Informatica PowerCentre           & 1993 & C & IPL   &   &   & B & B &   &   &   &   &   &   & B &   & B & B & I \\
    IBM InfoSphere DataStage          & 1996 & C & IDL   &   & B & B &   &   &   &   &   &   &   & B &   & B &   & B \\
    Microsoft SSIS                    & 2005 & C & SQL, .NET &   &   & B &   &   &   &   &   &   &   & B &   & B & B &  \\
    SAS Data Integration Studio       & 2005 & C & SAS   & B &   &   &   &   &   &   &   &   &   & B &   & B & B &   \\
    Oracle Data Integrator (ODI)      & 1996 & C & ODI Language &  & & B &   &   &   &   &   &   &   & B & B & B & B &  \\
    Qlik Replicate (AR)               &      & C &       &   &   & B &   &   &   &   &   &   &   &   & B & B & B &    \\
    SnapLogic                         & 2006 & C & Py, Jv, Ruby, JS & & & B & & & &  &   &   &   &   &   & B &   & B \\
    Xplenty                           & 2011 & C & SQL, Python, JS &  & & B & & & &  &   &   &   &   &   & B & B &   \\
    Matillion                         & 2011 & C & Python &   &   & B       & & & &  &   &   &   &   &   & B &   & B \\
    Striim                            & 2012 & C &SQL, XML,&  &   & B       & & & &  &   &   &   &   &   & B &   &  \\
                                      &      &   &JSON, JS &  &   &         & & & &  &   &   &   &   &   &   &   &  
    \\
    
   \hline
   
   \end{tabular}
   \\
   
   a. Microsoft SSIS:Microsoft SQL Server Integration Services; (AR): (formerly Attunity Replicate)\\
   b. Py: Python; IPL: Informatica PowerCenter language; JS: JavaScript; IDL: InfoSphere DataStage Language;
   Sc: Scala; Jv: Java
   \\Within the capabilities, B: Built-in capabilities; I: Integrates with other tools to achieve.

\end{table*}

\subsection{Orchestration and Workflow Management}
Table \ref{tab:Orch_Wkfl_Pipelines} presents Pipeline Orchestration and Workflow Management tools, frameworks, or platforms which are designed to centralise the automation, execution, monitoring, and management of workflow through the entire data pipelines from data sourcing through to utilization. These pipelines ensure that other interconnected pipelines or phases of a pipeline are executed in an orderly fashion to process data end-to-end.

\textit{Apache Airflow} is an open-source web-interfaced data pipeline orchestration and workflow management tool for ETL and data integration. It supports Python language and can perform automated workflow orchestration, scheduling, and monitoring of processes in parallel or as dependent processing of tasks and displays processes with graphs or DAGs to indicate the position and status of data processing (https://airflow.apache.org/). It works as a standalone or deployed to the cloud and performs data processing in batches but can integrate with frameworks like Apache Kafka to provide streaming data processing. Airflow is extensible and can integrate with other tools and API for increased capability, although has a steep learning curve for initial setup and operation.

\textit{Apache Beam} (https://beam.apache.org/) is open source and supports Python, Java, SQL, and GO and executes parallel processing, plus distributed processing for big data pipelines. 

\textit{Microsoft Azure Data Factory} (https://azure.microsoft.com) is a close source platform that runs on Azure-specific languages and provides a cloud-based service for data pipeline orchestration and workflow management.

\begin{table*}

    \renewcommand{\arraystretch}{1.0}
    \setlength{\tabcolsep}{3pt}

    \smallskip
    \caption{Orchestration and Workflow Management Pipelines}
    \label{tab:Orch_Wkfl_Pipelines}
    \begin{tabular}{|l|c|c|c|*{4}{p{0.3cm}|}*{4}{p{0.3cm}|}c|*{3}{p{0.3cm}|}c|*{2}{p{0.3cm}|}l|}
    \toprule
    \multirow{2}{*}{Tools/Framework/Platform} &
    \multicolumn{1}{c|}{\rotatebox{90}{Year of Introduction}} &  
    \multicolumn{1}{c|}{\rotatebox{90}{Source}} &
    \multicolumn{1}{c|}{\rotatebox{90}{Primary Language}} &
    \multicolumn{4}{c|}{\rotatebox{90}{Data Organisation}} & 
    \multicolumn{4}{c|}{\rotatebox{90}{Data Quality}} &
    \multicolumn{1}{c|}{} &
    \multicolumn{3}{c|}{\rotatebox{90}{Processing Format}} &
    \multicolumn{1}{c|}{} &
    \multicolumn{2}{c|}{\rotatebox{90}{Environment}} \\
    \cline{5-18}
    &&&& \rotatebox{90}{Data Parsing} & \rotatebox{90}{Data Dictionary} & \rotatebox{90}{Data Integration} & \rotatebox{90}{Data Transformation} & \rotatebox{90}{Canonicalization} & \rotatebox{90}{Missing Data} & \rotatebox{90}{Anomalies} & \rotatebox{90}{Non-Stationarity} & \rotatebox{90}{Feature Engineering} & \rotatebox{90}{ETL} & \rotatebox{90}{Batch} & \rotatebox{90}{Streaming} & \rotatebox{90}{DAG / Graph} & \rotatebox{90}{Local} & \rotatebox{90}{Cloud} \\
    \hline
    \\
    Apache Airflow	    & 2014 & O & Python	         &   & I & I & I & I & I & I & I & I & B & B & I & B & B & B \\
    Apache Beam	        & 2016 & O & Py, Jv, SQL, GO & I & I & B & I & I & I & I & I & I & B & B & B & B & B & B \\
    Microsoft Azure Data Factory	& 2015 & C & AL  & B &   &   &   &   &   &   &   &   &   & B & I & B &   & B \\
    Zapier	& 2011 & C & JavaScript				     &   &   &   &   &   &   &   &   &   &   &   &   & B &   & B \\
    \\
    
   \hline
   
   \end{tabular}
   \\
   
   a. Py: Python; Jv: Java, AL: Azure-specific Languages
   \\Within the capabilities, B: Built-in capabilities; I: Integrates with other tools to achieve.

\end{table*}

\begin{table*}[!htbp]

    \renewcommand{\arraystretch}{1.0}
    \setlength{\tabcolsep}{3pt}

    \smallskip
    \caption{Machine Learning and Model Deployment Pipelines}
    \label{tab:ML_MD_Pipelines}
    \begin{tabular}{|l|c|c|c|*{4}{p{0.3cm}|}*{4}{p{0.3cm}|}c|*{3}{p{0.3cm}|}c|*{2}{p{0.3cm}|}l|}
    \toprule
    \multirow{2}{*}{Tools/Framework/Platform} &
    \multicolumn{1}{c|}{\rotatebox{90}{Year of Introduction}} &  
    \multicolumn{1}{c|}{\rotatebox{90}{Source}} &
    \multicolumn{1}{c|}{\rotatebox{90}{Primary Language}} &
    \multicolumn{4}{c|}{\rotatebox{90}{Data Organisation}} & 
    \multicolumn{4}{c|}{\rotatebox{90}{Data Quality}} &
    \multicolumn{1}{c|}{} &
    \multicolumn{3}{c|}{\rotatebox{90}{Processing Format}} &
    \multicolumn{1}{c|}{} &
    \multicolumn{2}{c|}{\rotatebox{90}{Environment}} \\
    \cline{5-18}
    &&&& \rotatebox{90}{Data Parsing} & \rotatebox{90}{Data Dictionary} & \rotatebox{90}{Data Integration} & \rotatebox{90}{Data Transformation} & \rotatebox{90}{Canonicalization} & \rotatebox{90}{Missing Data} & \rotatebox{90}{Anomalies} & \rotatebox{90}{Non-Stationarity} & \rotatebox{90}{Feature Engineering} & \rotatebox{90}{ETL} & \rotatebox{90}{Batch} & \rotatebox{90}{Streaming} & \rotatebox{90}{DAG / Graph} & \rotatebox{90}{Local} & \rotatebox{90}{Cloud} \\
    \hline
    \\
    TFX   (TensorFlow Extended)  & 2017 & O & Python       & & & & & & &I&I&B& & &I&I& &B\\
    Kubeflow                     & 2017 & O & Py, C++, JS  & & & & & & & & & & & &I&I& &B\\
    Metaflow                     & 2018 & O & Python       & & & & & & & & & & &B&I&B& &B\\
    ZenML                        & 2020 & O & Python       & & & & & & & & & & & & &B& &B\\
    MLRun                        & 2019 & O & Python       & & & & & & & & & & & &I&I& &B\\
    CML                          &      & O & Python       & & &B& & & & & & & & & &I& &B\\
    Cortex Lab                   &      & O & Python       & & & & & & & & & & & & &B& &B\\
    Seldon Core                  & 2016 & O & Python, Go   & & & & & & & & & & & &I&I& &B\\
    AutoKeras                    & 2019 & O & Python       & & & & & & & & &I& & & &I& &B\\
    H2O AutoML                   & 2012 & O & Python, R    & & & & & & &I& &I& & & &I& &B\\
    \\
        
    \hline
   
   \end{tabular}
   \\
   
   a. Py: Python; JS: JavaScript;
   \\Within the capabilities, B: Built-in capabilities; I: Integrates with other tools to achieve.

\end{table*}

\subsection{Machine Learning and Model Deployment}
The output from a data engineering pipeline is consumed in a number of other processes; one such is an input for a machine learning pipeline. Machine learning pipelines some of which are listed in Table \ref{tab:ML_MD_Pipelines}, optimise input data and efficiently process ingested data through iterative model development, model training, model deployment, and monitoring. It is difficult to implement automatic machine learning (AutoML) throughout the entire iterative process for a given dataset \cite{De_Bie_2022}. One such difficulty is that it requires sophistication and experience to achieve. Like with pipelines that can be used for data engineering, orchestration is important in ML pipelines to ensure all parts of the pipeline work to prevent failure in the sequence of the processes. Optimisation on its part helps to ensure that the model works as built, while continuous monitoring ensures model drift is easily detected. The ideal case where possible is to automatically manage the overall ML pipeline from dataset to model deployment. Examples of ML tools/frameworks/pipelines for machine learning and model deployment are \textit{TFX (TensorFlow Extended)} and \textit{AutoKeras}.

\textit{TFX (TensorFlow Extended)} is an open-source machine learning pipeline platform from Google, it is based on TensorFlow and supports Python language, and can connect with API backends of C++, JavaScript, Java (https://www.tensorflow.org/tfx/guide). TFX provides a basis for building end-to-end ML pipelines by combining its standard components in sequential order and can be customised with custom components. It is extensible with inherent libraries, scalable to meet machine learning tasks, and integrates with other tools such as Apache Airflow and Kubeflow to provide orchestration functionality \cite{ru2020machine}. 

\textit{AutoKeras} is an open-source, AutoML framework based on Keras and compatible with Python and TensorFlow (https://autokeras.com/). Raw data processing, ML model building, deployment, optimisation, and management is automated and streamlined using AutoKeras' automated hyperparameter tuning. \textit{AutoKeras} is useful for automated deep-learning model selection, it's flexible, and extensible having built-in visualization to simplify deep learning for structured data, texts, and images \cite{jin2023autokeras}.

\section{Case Study with IDEAL Household Energy Dataset}
\label{Sec5}

The IDEAL Household Energy Dataset is a dataset collected from 255 homes in the United Kingdom covering 23 months and comprises sensor readings of electricity and gas, and other supplementary readings and documentation to add context to the dataset \cite{pullinger2021ideal}. The actual data is in a web location (https://datashare.ed.ac.uk /handle/10283/3647) and comprises sensor readings which are timestamps and values contained within .csv files zipped as .gz and finally as .zip. The filenames for each sensor reading consist of identifiers for the homes, rooms, sensor, sensortype, and types of readings, and where applicable these can be supplemented with additional information from the metadata records. There are over 16000 different .csv files and analysis of the data shows a number of data wrangling issues such as missing data, anomaly readings, different spellings of features, and missing primary keys to join tables. 

\subsection{Evaluating some framework tools}
The IDEAL Household Energy Dataset was reviewed using one example of each category of the tools/frameworks/platforms to implement a pipeline to demonstrate the possibilities of resolving some data engineering issues. Depending on the tool and tasks to accomplish, a sample household sensor dataset and metadata are used within the several IDEAL Household Energy Datasets available. The reason is that different issues are present in different sets of the data. So different samples are required to demonstrate the application of the chosen tools. The overall method is to \textit{download and unzip .zip or .gz the data}, next is to \textit{view the data structure}. Followed by some data engineering tasks in an attempt to perform some aspects of data organisation, data quality, and feature extraction to improve the data and get it ready for machine learning utilisation. Within the broad groups, the following tools were reviewed ETL/ELT Data Pipelines: \textit{Apache Spark with PySpark in Jupyter notebook}; Integration, Ingestion and Transformation: \textit{Microsoft SQL Server Integration Services}; Orchestration and Workflow Management: \textit{Apache Airflow}; Machine Learning and Model Deployment: \textit{TFX (TensorFlow Extended)}. These tools were selected from the broad groups based on their primary purpose of design, capabilities to resolve data wrangling issues, ability to allow for custom-built additions, levels of integration with other tools, levels of sophistication and adaptability, usability, and popularity within the user community.

\subsection{Pipeline with Apache Spark}
\begin{figure*}[h]
    \centering
    \raggedright 
    \includegraphics[width=16cm, height=3.75cm]{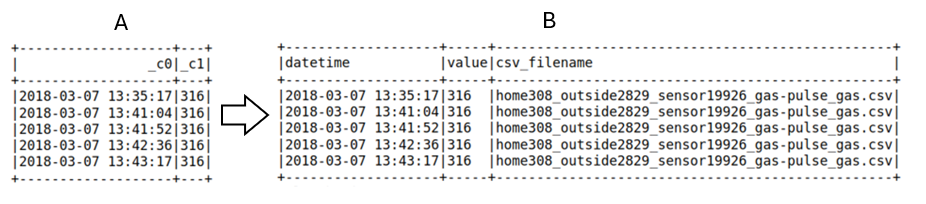}
    \caption{An indication of some data organisation, data
integration and feature engineering capabilities of Apache Spark. In \textbf{A} the initial reading of the household sensor csv data with spark.read.csv shows the absence of feature names. In \textbf{B} \textit{StructType} and \textit{StructField} were used to create dataframe schema and feature names, while the data types where assigned with \textit{TimestampType()} and \textit{IntegerType()}. The \textit{withColumn()} method was used to create a new feature csv\_filename populated with the source csv filenames obtained from the use of \textit{regexp\_extract} and \textit{input\_file\_name()}. Finally, these dataframes were integrated by \textit{.union}.}
    \label{fig:Apache_Spark_pipeline_nfn}
\end{figure*}

\textit{Apache Spark} environment was set up in Jupyter Notebook with PySpark, the Python API for Apache Spark. A Spark session was established with SparkSession builder to initialise the Spark session. For \textit{data parsing}, IDEAL Household Energy Dataset metadata and household\_sensors data were downloaded from the web with python library \textit{requests}. The household\_sensors is over 15GB zip file and takes several hours to download and process. As a result, a sample of the data is used as an example for the pipeline process. The metadata\_and\_survey.zip dataset was unzipped and extracted with \textit{zipfile}, while the household\_sensors.csv.gz was unzipped with \textit{zipfile} to access the .gz file and later unzipped with python library \textit{gzip}. The resultant content .csv files were read with \textit{spark.read.csv} into spark dataframes. A peek at the data was obtained using the PySpark \textit{.show()} method. Some data organisation, data integration, and feature engineering were performed on the resulting household\_sensors.csv files, these household\_sensors.csv files did not have feature names (Figure \ref{fig:Apache_Spark_pipeline_nfn}). Feature names and data types were assigned with a combination of PySpark \textit{StructType}, \textit{StructField}, \textit{TimestampType}, and \textit{IntegerType} based on the accompanying data dictionary documentation in the web repository of the dataset. A new column was created using the method \textit{withColumn()}, the new column contains the filename of the source data .csv extracted using a combination of \textit{regexp}\_\textit{extract} and \textit{input}\_\textit{file}\_\textit{name}. Using an example of the household sensor reading for home 308, data integration was performed with PySpark \textit{.union()} to combine the individual .csv sensor readings from the home.

The \textit{.printSchema()} was used to print the schema and provide an understanding of the feature names and data types. Some data types were changed for timestamp features from string datatype to timestamp data type using a combination of \textit{withColumn} and \textit{to}\_\textit{timestamp}. Further to \textit{data dictionary} the Pandas methods \textit{toPandas} \textit{describe}, and \textit{transpose} were used to convert the Spark dataframe to Pandas dataframe and display some statistics of the dataframe.

\begin{figure*}[!htbp]
    \centering
    \raggedright 
    \includegraphics[width=15.5cm, height=5.0cm]{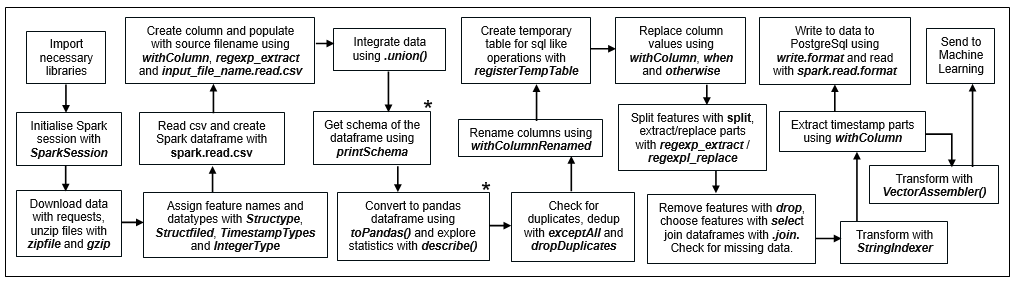}
    \caption{Apache Spark pipeline implemented to ingest and process the IDEAL Household Energy Dataset. The import of libraries and initiation of the Spark session prepared the pipeline before downloading the data. StructField and StrucType prepare the structure of the data in preparation for ingestion with spark.read.csv. Afterward, data transformation was performed using a combination of withColumn, regexp\_extract, and input\_file\_name. The resulting dataframes were integrated with .union to consolidate the data. The processes marked with asterisks help to view the data for better understanding without affecting the structure. Data cleaning was performed with dropDuplicates, columns renamed using withColumnRenamed, and column data altered using withColumn, when and otherwise. Some columns were split and parts were extracted. Dataframes were joined with .join, while the transformation of the data was with StringIndexer, fit, and transform to apply the index. Datetime parts were obtained before using appropriate JDBC connection and table names to write or read the dataframe into the PostgreSQL database. To write the dataframe with \textit{.write.format}, \textit{.mode,} \textit{.option} and \textit{.save} methods and later read the table using \textit{.read.format}, \textit{.option} and \textit{.load} methods. As machine learning input, the dataframe was transformed using \textit{VectorAssembler} to input the feature values into one single-named feature.}
    \label{fig:Apache_Spark_pipeline_flow_map}
\end{figure*}

Data deduplication checks and removal were performed with \textit{exceptAll} and \textit{dropDuplicates}. Further data transformation was performed in renaming column with PySpark method \textit{withColumnRenamed}. The data was explored with SQL-like operations by creating temporary tables using PySpark \textit{registerTempTable} and \textit{.sql()} to run SQL statements. For data transformation with PySpark the method \textit{withColumn}, \textit{when} and \textit{.otherwise} were used to replace data values within the columns. Some feature names were split with \textit{split} and part of the names extracted with \textit{regexp\_extract} and part of the names dropped with \textit{regexp\_replace}. Undesired features were dropped with \textit{.drop()}, and when required the desired features were obtained with \textit{select()}. Different dataframes were integrated across features with \textit{.join}, while the resultant dataframe was transformed using \textit{StringIndexer} to assign an index to categorical features, while \textit{.fit()} and \textit{.transform()} used to apply the index. On the datetime features, the parts of the datetime were extracted using \textit{dayofmonth}, \textit{month}, \textit{year}, \textit{hour}, \textit{minute}, \textit{seconds}. With an appropriate JDBC connection, the dataframe at this stage was written into PostgreSQL database using textit{.write.format}, \textit{.mode,} \textit{.option} and \textit{.save} methods. To read tables from the PostgreSQL database the methods \textit{.read.format}, \textit{.option} and \textit{.load} were used with the appropriate table names. To finalise as a machine learning input, the dataframe was transformed using \textit{VectorAssembler} to input the feature values into one single-named feature. A summary of the pipeline flow is illustrated in Figure \ref{fig:Apache_Spark_pipeline_flow_map}.

\subsection{Pipeline with Microsoft SSIS}
\textit{Microsoft SQL Server Integration Services (SSIS)} pipeline environment was set up in Visual Studio as an administrator to allow for bulk data load, and connected to a local server of Microsoft SQL Server. Microsoft SSIS is used for data integration and transformation within data organisation. It is common to use several built-in tasks and custom tasks within \textit{Control Flow} component of the SSIS graphical designer interface for designing workflows. 

\begin{figure}[!htbp]
    \raggedright 
    \begin{center}
    \includegraphics[width=7.75cm, height=9.5cm]{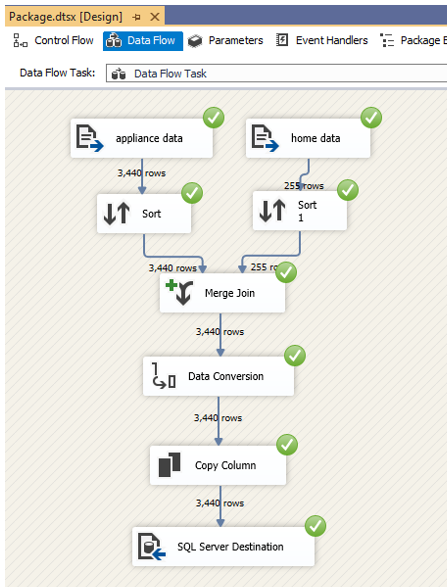}
    \end{center}
    \caption{Microsoft SSIS Pipeline shows the processing of the metadata of the IDEAL Household Energy dataset. Connection to the "appliance" and "home" data was established with SSIS Source Assistant. Both data were sorted and later integrated with the Merge Join component of SSIS. Some data transformation and standardization were achieved with SSIS Data Conversion before selecting some columns of interest with Copy Column, these were loaded into an SQL database with the SQL Server Destination component.}
    \label{fig:SSIS_Pipeline_1}
\end{figure}

For \textit{data parsing}, the \textit{Execute Process Task} which accepts and executes custom-designed codes was used to implement a Python code to download and unzip files to a local directory. The resulting .csv files were linked to a \textit{Data Flow Task} using \textit{Source Assistant} to configure connections to the example files, view some data dictionary elements, and preview the entire data structure. \textit{Sort} component was used to define sort order to enable data integration using \textit{Merge Join} component for a left join. The \textit{Data Conversion} component performed data transformation and standardization. The resultant data is linked with \textit{Copy Column} to select required columns before loading data into a local SQL server using \textit{SQL Server Destination} component as shown in Figure \ref{fig:SSIS_Pipeline_1}. 
Microsoft SSIS has a \textit{Data Cleansing} component within the data flow task for configuring data quality services used in data cleansing transformations. 

\subsection{Pipeline with Apache Airflow}
\textit{Pipeline with Apache Airflow} pipeline environment was configured to communicate with the PostgreSQL database which was managed using a SQL database 
administration tool DBeaver. The airflow database was initialised and the relevant airflow components were started first by running the airflow scheduler to schedule workflows and manage tasks, while the airflow webserver was started to view and manage airflow in a graphical web interface directed acyclic graphs (DAGs). 

\begin{figure*}[!htbp]
    \raggedright 
    \includegraphics[width=15.25cm, height=3.5cm]{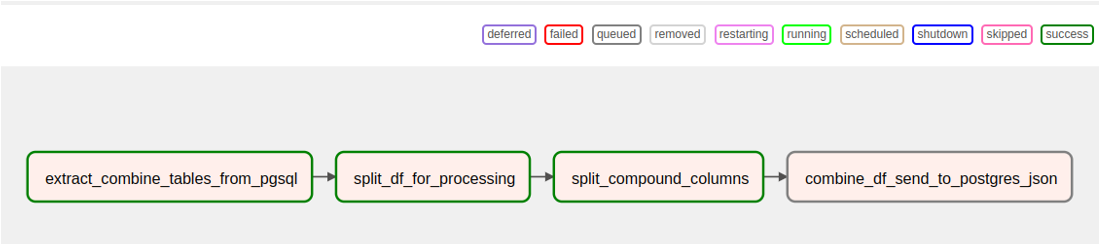}
    \caption{Airflow Pipeline from data ingestion, processing, and storage. The IDEAL Household Energy Dataset sensor data for different homes was already in a PostgreSQL database. The extract\_combine\_tables\_from\_pgsql task comprises Python codes to read different sensor data from homeid 308, slice the data across a time interval, create features, assign datatypes, integrate the data, and push the data to a processing location. The split\_df\_for\_processing task continues with the preprocessing of the data by performing some feature engineering. The split\_compound\_columns task processes the data by splitting, renaming, and reordering features. The combine\_df\_send\_to\_postgres\_json task combines the processed data and stores the data in a local repository as CSV and JSON, and saves the final dataset in a PostgreSQL database.}  
    \label{fig:airflow_pipeline}
\end{figure*}

Figure \ref{fig:airflow_pipeline} represents the Airflow DAG for processing the data in this example. The green outline tasks have completed a successful run while the grey outline task is scheduled to run next.

Apache Airflow tasks or dags files were written to schedule and sequentially organise airflow tasks, these tasks were run with a \textit{LocalExecutor}. A task was written in Python to download and extract the raw source .csv file, this task was called using Airflow \textit{PythonOperator} while \textit{PostgresHook} was used to connect with the PostgreSQL database to load the raw .csv files as tables. To clean the initial data, further tasks were written to read files off the database performing feature extraction as required. 

Tasks were automated and managed in sequential order by the in-built orchestration abilities of Apache Airflow. Before executing tasks, Airflow checks if the code is correct, for instance, it checks for references to variables, if a variable is non-existent then Airflow flags this as an issue to be fixed. In the event of failures, while executing dags Airflow provides a log pointing to the failure point and suggested causes. After fixing issues, dags can be executed to continue from the points where the failure occurred. 

\subsection{Pipeline with TFX (TensorFlow Extended)}
\begin{figure*}
    \raggedright 
    \includegraphics[width=15.6cm, height=4.75cm]{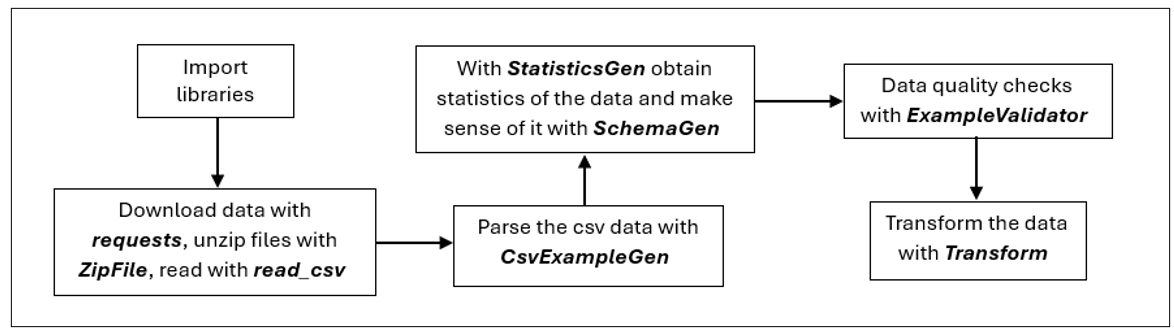}
    \caption{An Example TensorFlow Extended pipeline (adapted from https://www.tensorflow.org/tfx/guide) used for the case study. The required Python libraries were imported, with some required libraries for TensorFlow Extended such as TensorFlow and TFX. The specific TFX tasks were performed with the TFX components CsvExampleGen, StatisticsGen, SchemaGen, ExampleValidator, and Transform. The CsvExampleGen component is within ExampleGen and used to perform data ingestion, while StatisticsGen provides insight into the statistics of the data, SchemaGen as the term implies, generates data schema while the ExampleValidator is a data quality component for determining anomalies and missing values. The Transform component is required for feature extraction.}
    \label{fig:TFX_Extended_Pipeline}
\end{figure*}

\begin{figure*}
    \raggedright 
    \includegraphics[width=15.60cm, height=14.26cm]{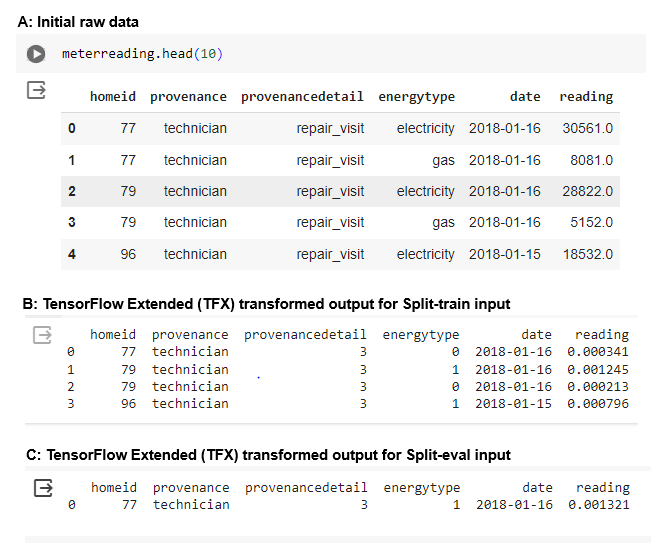}
    \caption{TensorFlow Extended (TFX) pipeline output after applying the TFX \textit{Transform} component. \\ \textbf{A} is the raw dataset, when parsed into TFX it splits into Split-train and Split-eval. \textbf{B} is the transformed Split-train and \textbf{C} the transformed Split-eval. Data transformation was applied to the numeric feature "reading" with TFX \textit{scale\_to\_0\_1} function, and the dimension of the result was reshaped with Tensorflow \textit{reshape} function to produce an output of values scaled between 0 and 1. For categorical features "provenancedetail" and "energytype", TFX \textit{compute\_and\_apply\_vocabulary} function was used to assign integer indices to the distinct dictionary values within these features. The most frequent values were assigned with indices 0 (for example "installation\_visit" in feature provenancedetail) while the least occurring values were assigned the maximum indices within the feature (for example "repair\_visit" is the least occurring and it is assigned value 3 within the feature provenancedetail).}
    \label{fig:TFX_raw_vs_transformed}
\end{figure*}

\textit{TFX (TensorFlow Extended)} pipeline environment was set up using the Google Colab interface to aid the availability of the required packages since TFX was built by Google. The pipeline involves downloading the data and aligning components to perform relevant tasks to resolve some issues with data organisation, data quality, and feature extraction to have data ready for machine learning. As illustrated in Figure \ref{fig:TFX_Extended_Pipeline} the example data was downloaded from the web location using python library \textit{requests} while the library \textit{zipfile} and \textit{gzip} were used to extract files respectively. 
The meterreading.csv dataset was the example file of interest, this was copied into a defined folder where the TFX component \textit{CsvExampleGen} of \textit{ExampleGen} was used to parse the CSV data and convert it into TFX format. In this example, the component splits the data into Split-train and Split-eval and subsequent components use these splits as inputs. Some data dictionary information where obtained with the \textit{StatisticsGen} used to obtain some statistics of the CSV, while the \textit{SchemaGen} creates a data schema. Data quality checks were performed with \textit{ExampleValidator} to define anomalies. The TFX \textit{Transform} component is used for feature extraction in preparation for training and analysing ML models (https://www.tensorflow.org/tfx/guide/transform). The transform component requires a python file which contains a custom script designed to perform a required transformation on the data. In Figure \ref{fig:TFX_raw_vs_transformed} a view of the raw meterreading.csv and the transformed output shows the effect of the TFX transformation applied. The values of a selected numeric feature were scaled from 0 to 1, while some selected categorical features had integer indices assigned to their distinct dictionary values. These indices were assigned from 0 to a maximum where 0 represents the most frequently occurring value. Like Airflow, orchestration occurs within TFX pipelines either locally or using Kubeflow when deploying pipelines in production (https://www.tensorflow.org/tfx/guide/local orchestrator).

\section{Discussion of Trends in Features}
\label{Sec6}
Generally, there are variations of available built-in components within pipelines useful for data engineering. These components can achieve certain levels of desired outcomes when processing data aimed at providing solutions to data organisation, data quality, and feature engineering. The components for these issues are sometimes updated to meet the demands of current or potential challenges with the use of data. Some pipeline tools also have customisation potentials for user communities to build tailored solutions to data engineering issues. For the tools considered against the data processed, there are still numerous components within each space that have not been utilised for the case study examples. 

Pipelines for Microsoft SQL Server Integration Services provide the advantage of having a brief description of what each task and component are used for, making it easier to understand potential outcomes before using the component. Setting up an SSIS environment requires some workaround from download, installation, configuration, and storage of the processed data. Failures along the pipeline can be addressed using some troubleshooting logs which are readily available to help address challenges, especially with the added advantage of its graphic user interface.

Apache Spark pipelines as PySpark in Jupiter Notebook appear to have an easier setup and processing while using available Python methods. Installation and getting ready, as well as data processing of the example, was a lot easier. Although, the programming aspect ensures a user must have the right codes unlike in SSIS where built-in graphical drag-and-drop solutions are sometimes available.

Apache Airflow data pipelines provided a robust data engineering solution for the case treated, however, Airflow had a steep learning curve. Initial setup, configuration, and integration with other tools were somewhat difficult for an uninformed user but afterward, the use as a data engineering pipeline provided a richer solution that can be custom-built to provide additional benefits as the pipeline evolves. Airflow's Python base and integration with multiple tools, available graphical interface DAG, and its web interface aided usability to accomplish set tasks. Generally, after the initial challenges of setting up, it was easier to accomplish tasks also managing, and monitoring workflows were more flexible. 

TensorFlow Extended (TFX) provides a pipeline for end-to-end machine learning from data ingestion and validation through to model training, analysis, and deployment. Setting up TFX with Google Colab was without difficulties, the library TensorFlow Data Validation (TFDV) provides scalable data analytics for data, produces a view of data statistics and anomalies within the data, and detects changes to the variation of feature values \cite{steidl2023pipeline}. These advantages from TFDV help to minimise input data errors and invariably helps with data quality in machine learning data handling \cite{caveness2020tensorflow}.

\section{Recommendation}
\label{Sec7}
The evaluation of the different tools informs the choice of Apache Airflow as the preferred option for implementing a data engineering pipeline. Apache Airflow is open source and it is highly scalable in terms of adding tasks over time. Customisation is achieved for tasks and workflows with easily defined Python codes and can link up with existing tasks along the pipeline. Airflow is an open container within which different data wrangling solutions can be implemented. Airflow has built-in and allows for customised hooks and operators. Hooks are used for connecting with external systems, while operators are required for executing tasks. This extensibility allows for a combination of task operations within Apache Airflow rather than in an individual operator's environment. For instance, instead of working within the Spark (PySpark) environment, Airflow has plug-ins such as the Apache Spark Operators, Apache Airflow Provider for Apache Spark, and PySpark Decorator that can all be used in Apache Airflow to run Apache Spark jobs and manage the workflows. Apache Airflow can be used to run TFX jobs as an orchestrator and workflow manager. While the initial setup for Airflow preferably in a Linux machine might be a steep learning curve or even more challenging in a Windows machine with or without Docker, Apache Airflow does however provide an environment for reproducing a code, monitoring, and managing operations. It can provide the option of running codes in isolated environments such that a local code testing environment can have a different setting and configuration from the production environment within the same Apache Airflow. This makes it more convenient to locally test a pipeline in isolation while configuring the production aspect of the pipeline.

\section{Conclusion}
\label{Sec8}
The most appropriate pipeline used for data engineering depends on the data, the tool, the expertise of using the tool, and the solution to provide. Essentially, more than one pipeline can be built at the exploratory phase of a solution to provide an understanding of the data and the adequacy of the solution to meet expectations. Although the Apache Airflow learning curve is steep at the initial stages of set and exploration, usability becomes easier, aided by a rich public resource, user community, and the Python framework. Also as the need arises to add more data to a workflow and maintain orchestration and management, Airflow appears easier to connect a DAG to the workflow and ensure continuous processing. While considering the overall benefits of Airflow as the preferred, it was easier to quickly set up and run Spark using Pyspark in Jupyter Notebook.

\begin{appendices}
\section{The Code}

As presented under Section \ref{Sec5}, some capabilities of selected pipelines were demonstrated on samples of the IDEAL Household Energy Dataset. The codes used for this evaluation have been saved in the GitHub repository: https://github.com/DataEng-AML/surveypaper. Reference pipeline in boldface and relevant code name listed:\\\\
    \textbf{Apache Spark with PySpark in Jupyter notebook} 
    \begin{itemize}
        \item ideal\_spark\_survey\_paper.ipynb
    \end{itemize}    
    \textbf{Apache Airflow}  
    \begin{itemize}
        \item replace\_dash\_x\_filename.py
        \item {ideal\_dataset\_processing.py}
    \end{itemize}  
    \textbf{TFX (TensorFlow Extended)}  
    \begin{itemize}
        \item TFX\_DE\_Survey\_paper.ipynb
    \end{itemize}   
\end{appendices}

\printbibliography
\end{document}